\documentclass[lettersize,journal]{IEEEtran}
\usepackage{amsmath,amsfonts}
\usepackage{algorithmic}
\usepackage{algorithm}
\usepackage{array}
\usepackage[caption=false,font=normalsize,labelfont=sf,textfont=sf]{subfig}
\usepackage{textcomp}
\usepackage{stfloats}
\usepackage{url}
\usepackage{verbatim}
\usepackage{graphicx}
\usepackage{cite}
\usepackage{xcolor}

\usepackage{multirow}
\usepackage{soul}
\usepackage{makecell}
\usepackage{wrapfig}
\usepackage{float}

\begin{document}

\title{A Mixed-Primitive-based Gaussian Splatting Method for Surface Reconstruction}

\author{Haoxuan~Qu,
        Yujun~Cai,
        Hossein~Rahmani,
        Ajay~Kumar,
        Junsong~Yuan,
        and~Jun~Liu
\thanks{H. Qu, H. Rahmani, and J. Liu are with Lancaster University, United Kingdom. Y. Cai is from The University of Queensland, Australia. A. Kumar is from The Hong Kong Polytechnic University, China. J. Yuan is from University at Buffalo, United States of America.\\
E-mail: h.qu5@lancaster.ac.uk, yujun.cai@uq.edu.au, h.rahmani@lancaster.a\\c.uk, ajay.kumar@polyu.edu.hk, jsyuan@
buffalo.edu, j.liu81@lancaster.ac.uk}
\thanks{Manuscript received April 19, 2021; revised August 16, 2021.\\(Corresponding author: Jun Liu.)}}

\markboth{Journal of \LaTeX\ Class Files,~Vol.~14, No.~8, August~2021}%
{Shell \MakeLowercase{\textit{et al.}}: A Sample Article Using IEEEtran.cls for IEEE Journals}


\maketitle

\begin{abstract}
Recently, Gaussian Splatting (GS) has received a lot of attention in surface reconstruction. However, while 3D objects can be of complex and diverse shapes in the real world, existing GS-based methods only limitedly use a single type of splatting primitive (Gaussian ellipse or Gaussian ellipsoid) to represent object surfaces during their reconstruction. In this paper, we highlight that this can be insufficient for object surfaces to be represented in high quality. Thus, we propose a novel framework that, for the first time, enables Gaussian Splatting to incorporate multiple types of (geometrical) primitives during its surface reconstruction process. Specifically, in our framework, we first propose a compositional splatting strategy, enabling the splatting and rendering of different types of primitives in the Gaussian Splatting pipeline. In addition, we also design our framework with a mixed-primitive-based initialization strategy and a vertex pruning mechanism to further promote its surface representation learning process to be well executed leveraging different types of primitives. Extensive experiments show the efficacy of our framework and its accurate surface reconstruction performance.  
\end{abstract}


\section{Introduction}
\label{sec:intro}

\IEEEPARstart{S}{urface} reconstruction aims to accurately reconstruct 3D object surfaces from multi-view RGB images. 
It is a fundamental task in 3D computer vision, and it is relevant to various applications, 
such as virtual reality \cite{deng2022fov} and content generation \cite{liu2024comprehensive}. 
In recent years, to perform accurate surface reconstruction, various Neural-Radiance-Field-based (NeRF-based) surface reconstruction methods have been proposed \cite{wang2021neus,li2023neuralangelo,fu2022geo}. 
Yet, these methods typically rely on a computationally intensive volume rendering scheme. 
This often leads these methods to have a long training time \cite{hyung2024effective}, hindering their usage in real-world applications.

More recently, thanks to its much shorter training time, Gaussian Splatting \cite{kerbl20233d} has been explored as an attractive alternative to NeRF, and many Gaussian-Splatting-based (GS-based) surface reconstruction methods have been proposed \cite{dai2024high,huang20242d,chen2024pgsr}. 
Specifically, while Gaussian Splatting originally represents the 3D scene with 3D Gaussian ellipsoids, in surface reconstruction, to better conform to object surfaces, many recent GS-based methods often instead equip Gaussian Splatting with planer Gaussian ellipses \cite{dai2024high,huang20242d,chen2024pgsr}.
By doing so, these GS-based methods combine the efficiency advantage of Gaussian Splatting and the surface alignment advantage of Gaussian ellipses, and have achieved good performance in surface reconstruction. 
The GS-based surface reconstruction methods have then received a great deal of research attention \cite{dai2024high,huang20242d,chen2024pgsr,guedon2024sugar}.  

\begin{figure}[t]
\centering
\includegraphics[width=\columnwidth]{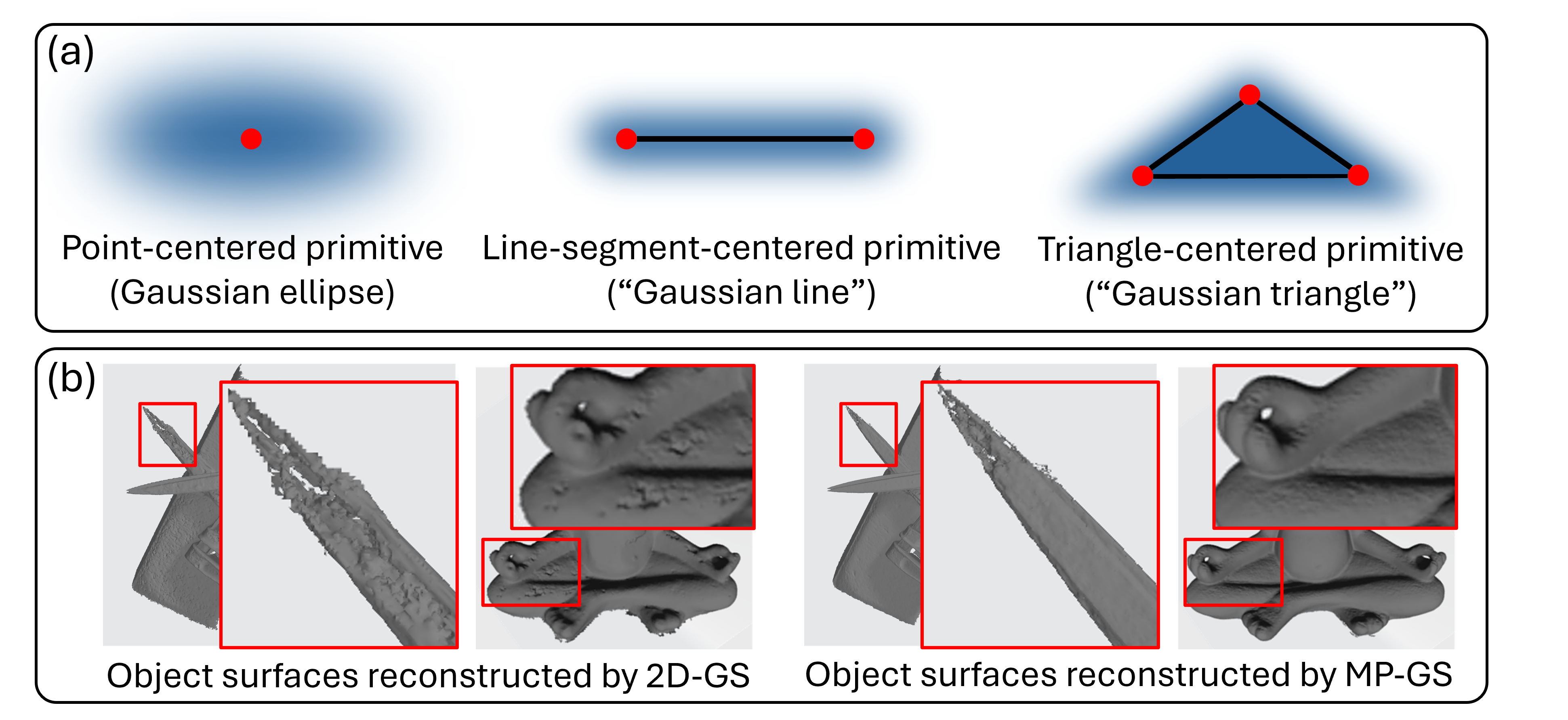}
\caption{(a) Illustration of the three different types of splatting primitives our framework uses. (b) Illustration of object surfaces reconstructed by 2D-GS \cite{huang20242d} and our MP-GS framework. As shown, only leveraging Gaussian ellipses as the primitive, 2D-GS can fail to accurately reconstruct object surfaces. In contrast, MP-GS, based on mixed types of primitives, enhances the reconstruction quality of object surfaces. \textbf{More qualitative results are in Fig.~\ref{fig:qualitative} and supplementary.} (Best viewed in color.)}
\label{fig:intro}
\end{figure}

Nevertheless, we argue that the GS-based methods solely utilizing planer Gaussian ellipses may still be sub-optimal in performing accurate surface reconstruction.
This is because, in real-life scenarios, objects can be of complex and diverse shapes. 
In this case, the Gaussian ellipse formulated by radial fading from a single point (as shown in Fig.~\ref{fig:intro}(a)), with its shape only limitedly controlled by its covariance, can fail to consistently well-represent the surfaces of different objects with different shapes \cite{pajarola2004confetti}. 
Indeed, as shown in Fig.~\ref{fig:intro}(b), for 2D-GS \cite{huang20242d} as a commonly used GS-based surface reconstruction method, only using 2D Gaussian ellipses, it can fail to reconstruct surfaces of different 3D objects in high quality. 
The above implies that in Gaussian Splatting, relying solely on point-centered Gaussian ellipses—and thus restricting the fading pattern to point-centered fading—can result in sub-optimal surface reconstruction performance.

In fact, as shown in previous reconstruction works predating the advent of Gaussian Splatting \cite{weber1995creation,deussen2002interactive,wong2005rendering}, relying solely on points as 0-dimensional simplices and formulating only point-centered primitives is often insufficient for accurately representing the shapes of many objects. Instead, line-segment-centered and triangle-centered primitives, respectively corresponding to line segments and triangles as 1-dimensional and 2-dimensional simplices, are also beneficial.
For example, as shown in \cite{weber1995creation,deussen2002interactive}, to accurately represent the shape of plants, especially over their sub-areas with long and thin structures such as their twigs, the line-segment-centered primitive is often regarded as a more fitting representation primitive compared to the point-centered one.
Meanwhile, as shown in \cite{wong2005rendering}, to accurately represent human joints such as the knee, rather than only utilizing point-centered representation primitives, additional utilization of line-segment-centered and triangle-centered primitives can generally lead to improved representation quality.

Inspired by the above, in addition to (point-centered) Gaussian ellipses, we aim to also design Gaussian Splatting with line-segment-centered and triangle-centered splatting primitives. As shown in Fig.~\ref{fig:intro}(a), these two types of primitives can respectively perform line-segment-centered and triangle-centered fading. Based on such design, Gaussian Splatting can then \textit{simultaneously utilize mixed types of primitives with different geometrical shapes and various fading patterns}, resulting in its more accurate reconstruction of object surfaces.

However, achieving the above goal can be non-trivial due to the following challenges:
(1) While the splatting of Gaussian ellipses can be straightforward following existing math formulas \cite{zwicker2002ewa}, these formulas only work for elliptical or ellipsoidal primitives. 
This leads the splatting of the other two non-elliptical primitives (i.e., the line-segment-centered and triangle-centered primitives) to be still challenging.
(2) Meanwhile, with the mixed types of primitives, how to utilize them together effectively in Gaussian Splatting is also difficult.
To handle the above challenges, in this work, we propose \textbf{M}ixed \textbf{P}rimitive-based \textbf{G}aussian \textbf{S}platting (\textbf{MP-GS}), a novel framework that for the first time, enables Gaussian Splatting to seamlessly incorporate both non-elliptical and elliptical primitives during its surface reconstruction process. 
By collaboratively using different types of splatting primitives with varying shapes and diverse fading patterns, MP-GS enables a more accurate reconstruction of object surfaces.
Below, we outline our MP-GS framework.   
Besides, in the rest of this work, for simplicity, we call the line-segment-centered primitives ``Gaussian lines'', and the triangle-centered primitives ``Gaussian triangles''.

Overall, to perform Gaussian Splatting based on mixed types of primitives, MP-GS first needs to enable the splatting of those newly-introduced non-elliptical primitives including ``Gaussian lines'' and ``Gaussian triangles''. 
To achieve this, we observe that, both the line segment and the triangle can be represented as the composition of their vertices. 
Inspired by this, in MP-GS, rather than performing splatting over the two non-elliptical primitives each as a whole which can be difficult, we instead propose a strategy to perform such splatting in a compositional manner via the following three steps. 
Specifically, given a viewpoint, to splat a ``Gaussian line'' or a ``Gaussian triangle'' onto its corresponding image plane, we first splat the primitive's ``vertices'' onto the image plane leveraging the well-established point-based splatting technique \cite{zwicker2002ewa,kerbl20233d}. 
Next, on the image plane, we re-sketch the ``Gaussian line'' from its two splatted ``vertices'' or the ``Gaussian triangle'' from its three splatted ``vertices''.
Finally, in MP-GS, we design a modified $\alpha$-blending function, by which image rendering can be performed over the re-sketched ``Gaussian lines'' and ``Gaussian triangles'', in a similar manner as the original Gaussian ellipses. 

Through the above process, we can successfully splat and render ``Gaussian lines'' and ``Gaussian triangles''. However, the above process alone does not fully support a mixed-primitive-based learning procedure for Gaussian Splatting. 
This is because, besides splatting and rendering as the key steps, the learning procedure of Gaussian Splatting also contains other steps. Among these steps, some of them, like the initialization and pruning steps, are also incompatible with the mixed-primitive-based nature of MP-GS. 
To tackle this issue, in MP-GS, we also propose two other designs: a mixed-primitive-based initialization strategy and a vertex pruning mechanism. By integrating these designs, our MP-GS framework finally enables Gaussian Splatting to be seamlessly and effectively performed in a mixed-primitive-based manner, allowing for more accurate representation and reconstruction of object surfaces.

The contributions of our work are as follows.
1) We propose MP-GS, a novel framework for surface reconstruction. To the best of our knowledge, this is the first effort that enables Gaussian Splatting to perform splatting based on mixed types of primitives during its surface reconstruction process.
2) We introduce several designs in MP-GS to enable the splatting and rendering of non-elliptical primitives, while also to facilitate the effective execution of the other steps in Gaussian Splatting in a mixed-primitive-based manner.
3) MP-GS achieves superior performance on the evaluated benchmarks.

\section{Related Work}
\label{sec:related}

\noindent\textbf{Surface reconstruction} has garnered significant research attention \cite{furukawa2009accurate,schonberger2016pixelwise,wang2021neus,fu2022geo,li2023neuralangelo,huang20242d,yu2024gaussian,fan2024trim,zhang2024rade,chen2024pgsr,dai2024high,wolf2024gs2mesh,lyu20243dgsr,chen2023neusg,wu2025surface,huang2024sur2f,jiang2024rethinking,wang2024neurodin,yu20242dgh,Jiang_2025_CVPR,wu2025sparse2dgs,Peng_2025_CVPR,toussaint2025probesdf,Tan_2025_CVPR,zhang2024quadratic,wang2024gaussurf,li2025monogsdfexploringmonoculargeometric,li2025tsgs} due to its extensive real-world applications. 
Initially, this task was primarily tackled using multi-view stereo techniques, broadly classified into depth map estimation and merging \cite{schonberger2016pixelwise,liu2009continuous}, voxel grid optimization \cite{seitz1999photorealistic,sinha2007multi}, and feature point growing methods \cite{furukawa2009accurate,wu2010quasi}. Over time, the exploration of neural rendering \cite{tewari2020state} has gained momentum, and many NeRF-based surface reconstruction methods were then developed, such as NeuS \cite{wang2021neus}, Geo-NeuS \cite{fu2022geo}, and Neuralangelo \cite{li2023neuralangelo}. Despite the increased effort, a key weakness of NeRF-based methods can be that, to perform accurate surface reconstruction, they generally demand a computationally heavy volume rendering procedure. 
This can suffer these methods from a long training time \cite{chen2024pgsr,hyung2024effective}, and thus limit their usage in many real-life scenarios.

In light of this, more recently, motivated by the high efficiency of Gaussian Splatting, many GS-based surface reconstruction methods have been proposed. Huang et al. \cite{huang20242d} replaced 3D Gaussian ellipsoids with 2D Gaussian ellipses and introduced a ray-splat intersection scheme for perspective-accurate splatting. Yu et al. \cite{yu2024gaussian} formulated a Gaussian Opacity Field, enabling direct surface extraction via identifying the level-set of the formulated field. Later, Chen et al. \cite{chen2024pgsr} proposed to further enhance planar-based Gaussian Splatting with techniques including unbiased depth rendering and single/multi-view regularization. 

Existing GS-based surface reconstruction methods typically rely on a single type of primitive—either planar Gaussian ellipses or 3D Gaussian ellipsoids. Likewise, GS-based methods in other tasks \cite{kerbl20233d,hamdi2024ges,zhang2024fregs,huang2025deformable,zhu20253d,arunan2025darbsplattinggeneralizingsplattingdecaying,tang2024dreamgaussian,qu2024disc} generally adhere to this convention, using only primitives with point-centered fading patterns, such as Gaussian ellipsoids, to represent the 3D scene. Differently, in this work, for the first time, we propose a GS-based surface reconstruction framework that supports mixed types of (elliptical and non-elliptical) primitives with diverse fading patterns and geometric shapes.

\section{Background on Gaussian Splatting}

Gaussian Splatting explicitly represents the 3D scene (object) as a set of Gaussian distributions. 
In specific, in the set, Gaussian Splatting defines each Gaussian with the following properties: (1) its center point $\mu \in \mathbb{R}^3$, (2) its covariance matrix $\Sigma \in \mathbb{R}^{3 \times 3}$, (3) its opacity $\alpha \in \mathbb{R}^1$, and (4) its spherical harmonic (SH) coefficients $c_{SH} \in \mathbb{R}^{3 \times (k+1)^2}$ representing its view-dependent color, where $k$ denotes the order of SH. 
Notably, to keep the covariance matrix $\Sigma$ positive semi-definite throughout learning, Gaussian Splatting further expresses $\Sigma$ as $\Sigma = RSS^TR^T$, where $R \in \mathbb{R}^{3 \times 3}$ and $S \in \mathbb{R}^{3 \times 3}$ respectively are the orthogonal rotation matrix and the diagonal scale matrix of the Gaussian.

With each Gaussian in the set defined in the above way, to perform image rendering over a given viewpoint, Gaussian Splatting first splats (projects) each Gaussian in the set onto the image plane corresponding to the viewpoint following the formulas in \cite{zwicker2002ewa} as:
\begin{equation}\label{eq:pre_1}
\begin{aligned}
\mu^{2D} = (PW\mu)[:2], ~ \Sigma^{2D} = (JW\Sigma W^TJ^T)[:2, :2]
\end{aligned}
\end{equation}
where $\mu^{2D} \in \mathbb{R}^2$ is the center point of the projected Gaussian distribution, $\Sigma^{2D} \in \mathbb{R}^{2 \times 2}$ is the covariance matrix of the projected Gaussian distribution, $W$ is the viewing transformation matrix, $P$ is the projective transformation matrix, and $J$ is the Jacobian of the affine approximation of the projective transformation. After that, to perform rendering on the image plane, Gaussian Splatting conducts $\alpha$-blending. Specifically, taking the rendering of the RGB image as an example, for each pixel $p$ of the image, Gaussian Splatting renders its RGB color $C(p)$ through $\alpha$-blending as:
\begin{equation}\label{eq:pre_2}
\begin{aligned}
& C(p) = \sum_{i=1}^M c_i \gamma_i \prod_{j=1}^{i-1}(1 - \gamma_j), \\
& \textbf{where}~ \gamma_i = \alpha_i e^{-\frac{1}{2}(p - \mu^{2D}_i)^T(\Sigma^{2D}_i)^{-1}(p - \mu^{2D}_i))}
\end{aligned}
\end{equation}
where $M$ is the number of projected Gaussians that overlap the pixel $p$, $c_i$ is the color of the $i$-th Gaussian calculated from the Gaussian's SH coefficient, $\alpha_i$ is the opacity of the $i$-th Gaussian, and $\mu^{2D}_i$ and $\Sigma^{2D}_i$ respectively denote the center point and the covariance matrix of the $i$-th projected Gaussian. 
Notably, the $\alpha$-blending function in Eq.~\ref{eq:pre_2} can be used for more than rendering RGB images. In fact, simply via replacing $c_i$ in Eq.~\ref{eq:pre_2} with other characteristics of the Gaussian, the function can also be used to render other types of images such as the depth map \cite{chen2024pgsr,jiang2024gaussianshader}. Meanwhile, also note that, no matter using 3D Gaussian ellipsoids or 2D Gaussian ellipses to represent the 3D scene (object), Gaussian Splatting can consistently perform rendering using the above equations. Indeed, as mentioned in existing surface reconstruction works \cite{huang20242d,dai2024high}, to perform Gaussian Splatting leveraging 2D Gaussian ellipses and thus enable them to better conform to object surfaces, a simple way (that we also follow in this work) is to just fix the last column of the scale matrix $S$ for each Gaussian to be a zero column vector.

\section{Proposed Method}

Given a batch of images of a 3D object along with their corresponding viewpoints, surface reconstruction aims to reconstruct the object's surface.
To handle this task, recently, Gaussian-Splatting-based (GS-based) methods, due to their accuracy and fast training speed, have attracted lots of research attention \cite{dai2024high,huang20242d,chen2024pgsr}. Yet, we here argue that existing GS-based methods can still result in sub-optimal representations of object surfaces, as shown in Fig.~\ref{fig:intro}(b). 
This is because, real-world 3D objects can be of complex and diverse shapes. However, existing GS-based methods typically rely on only a single type of primitive (e.g., the Gaussian ellipse) to represent object surfaces. This may be insufficient for capturing the full complexity of diverse object shapes. 

To tackle this problem, in this work, we propose a novel MP-GS framework, which can for the first time, enable Gaussian Splatting to represent object surfaces by using mixed types of primitives including Gaussian ellipses, ``Gaussian lines'', and ``Gaussian triangles'' collaboratively. 
Specifically, in MP-GS, we first propose a compositional splatting strategy (as described in Sec.~\ref{sec:4_1}). Leveraging this strategy, we enable the splatting of the non-elliptical primitives including ``Gaussian lines'' and ``Gaussian triangles'', and correspondingly allow the rendering of these primitives. 
After that, to further promote the learning procedure of MP-GS to be well executed in a mixed-primitive-based manner, we propose two additional adjustments to the typical Gaussian splatting pipeline in MP-GS, respectively editing its initialization and pruning steps (as introduced in Sec.~\ref{sec:4_2}).

\subsection{Compositional Splatting Strategy}
\label{sec:4_1}

To perform mixed-primitive-based Gaussian Splatting in the proposed MP-GS. the pre-requisite is to enable the splatting of the newly introduced non-elliptical primitives, including ``Gaussian lines'' and ``Gaussian triangles''. Yet, unlike the splatting of Gaussian ellipses, which can be done following established math formulas \cite{zwicker2002ewa}, the splatting of either “Gaussian line” or “Gaussian triangle” as a whole, with no existing splatting formulas, presents a challenge. 
To address this challenge, inspired by that both line segments and triangles can be regarded as the composition of their vertices, in MP-GS, we propose a compositional splatting strategy for enabling the splatting and subsequent rendering of ``Gaussian lines'' and ``Gaussian triangles''. 
To ease understanding, below, we first explain how the (compositional) splatting and rendering of ``Gaussian triangles'' is performed in MP-GS. 
We then discuss the splatting and rendering of ``Gaussian lines'' at the end of this section. Specifically, to compositional splat and render a ``Gaussian triangle'', we modify the typical Gaussian Splatting procedure of Gaussian ellipses via the following four steps.

\textbf{Step (1): Primitive formulation (definition).} Firstly, different from the Gaussian ellipse which only has a single center point, the ``Gaussian triangle'' has three distinct ``vertices''. This makes defining a ``Gaussian triangle'' also only with a single center point $\mu$ to be inadequate. Considering this, to properly define a ``Gaussian triangle'', we replace the original center point property $\mu$ of the Gaussian ellipse with a set of three parameters including $\mu_1 \in \mathbb{R}^3$, $\mu_2 \in \mathbb{R}^2$, and $\mu_3 \in \mathbb{R}^2$. Specifically here, for the first ``vertex'' of the ``Gaussian triangle'', we store its 3D coordinate directly in $\mu_1$. Yet, for the second and third ``vertices'', we leverage the fact that all three vertices of a triangle lie on the same plane, which, in our ``Gaussian triangle'' case, can be fully determined by $\mu_1$ and the rotation matrix $R$ of the ``Gaussian triangle''. Exploiting this property, to avoid redundancy, we store the second and third ``vertices'' in $\mathbb{R}^2$ using $\mu_2$ and $\mu_3$ rather than using full 3D coordinates. Notably, with $\mu_1$ and $R$, we can reconstruct the 3D coordinates of the second and third ``vertices'' $\mu_2^{3D} \in \mathbb{R}^3$ and $\mu_3^{3D} \in \mathbb{R}^3$ from $\mu_2$ and $\mu_3$ simply as follows:

\begin{equation}\label{eq:method_1}
\begin{aligned}
& \mu_2^{3D} =  \mu_1 + \mu_2[0] \times (R[:, 0])^T + \mu_2[1] \times (R[:, 1])^T \\
& \mu_3^{3D} =  \mu_1 + \mu_3[0] \times (R[:, 0])^T + \mu_3[1] \times (R[:, 1])^T
\end{aligned}
\end{equation}

For the rest properties of the ``Gaussian triangle'' primitive—including the covariance matrix $\Sigma$, the SH coefficients $c_{SH}$, and the opacity $\alpha$—we retain the same definitions as those well-established for Gaussian ellipses. Notably, for the covariance matrix $\Sigma$, here to reduce parameter overhead, we only store a single copy of it for the ``Gaussian triangle'', and we let all three vertices of the triangle share the same covariance matrix. With the properties of the ``Gaussian triangle'' properly defined in the above way, we describe how we perform splatting and rendering over it below.

\textbf{Step (2): Compositional splatting.} Given a viewpoint, we aim to splat each ``Gaussian triangle'' onto the viewpoint's image plane, feasibly in a compositional manner.
To achieve this, we point out that, it is enough to splat the ``vertices'' of the ``Gaussian triangle'' alongside their covariance matrix onto the image plane in a way similar to Eq.~\ref{eq:pre_1} as:
\begin{equation}\label{eq:method_2}
\begin{aligned}
& \mu_1^{2D} = (PW\mu_1)[:2], ~ \mu_2^{2D} = (PW\mu_2^{3D})[:2], \\
& \mu_3^{2D} = (PW\mu_3^{3D})[:2], ~ \Sigma^{2D} = (JW\Sigma W^TJ^T)[:2, :2]
\end{aligned}
\end{equation}
With these elements splatted, below, we discuss how they can be used to properly re-sketch and render the ``Gaussian triangle'' on the image plane.

\textbf{Step (3): Re-sketching on the image plane.} After obtaining $\mu_1^{2D}$, $\mu_2^{2D}$, $\mu_3^{2D}$, and $\Sigma^{2D}$, here, we aim to use them to re-sketch the boundary of the ``Gaussian triangle'' on the image plane. As shown in Fig.~\ref{fig:method} from (a) to (c), this is achieved in two sub-steps, via first formulating the boundary ellipses of the ``Gaussian triangle'' in step (3.1), followed by pairwise connecting these ellipses and finally deriving the boundary of the ``Gaussian triangle'' in step (3.2). 

\ul{Step (3.1): Boundary ellipse formulation.} Here, we first formulate the boundary ellipse centered at each ``vertex'' of the ``Gaussian triangle'' (i.e., the yellow ellipse centered at each ``vertex'' in Fig.~\ref{fig:method}(b)). 
In specific, we define the boundary ellipse as the ellipse whose contour is formulated by the set of points with Gaussian value to be $\frac{1}{255}$. 
According to Gaussian Splatting \cite{kerbl20233d}, to avoid numerical instability, the Gaussian values of points outside this ellipse are truncated to zero. 
Thus, the contour of this boundary ellipse can be regarded as the ``boundary'' of its corresponding Gaussian distribution in Gaussian Splatting.
With the three boundary ellipses of the ``Gaussian triangle'' acquired here in step (3.1), we can then use parts of their contours below in step (3.2) to form parts of the boundary of the ``Gaussian triangle'' (i.e., the solid-red-line part in Fig.~\ref{fig:method}(c)). 

Specifically, to form the contour of the boundary ellipse centered at the $i$-th ``vertex'' of the ``Gaussian triangle'', denoting $\mu_i^{2D} = (x_{\mu_i}, y_{\mu_i})$ and $\Sigma^{2D}$ as a symmetric matrix to be $\Sigma^{2D} = 
\begin{pmatrix} 
a_{\Sigma} & b_{\Sigma} \\
b_{\Sigma} & c_{\Sigma}
\end{pmatrix}$, we have the contour formed as the following (with the derivation provided in supplementary):
\begin{equation}\label{eq:method_3}
\begin{aligned}
& \{(x, y) | g^{ct}_i = 0\}, \textbf{where}~g^{ct}_i = \Big(-2\ln255~+ \\
& \frac{c_{\Sigma}(x_{\mu_i}-x)^2 + a_{\Sigma}(y_{\mu_i}-y)^2 + b_{\Sigma}(x_{\mu_i}-x)(y_{\mu_i}-y)}{a_{\Sigma}c_{\Sigma} - (b_{\Sigma})^2} \Big) 
\end{aligned}
\end{equation}
Via Eq.~\ref{eq:method_3}, given a ``Gaussian triangle'', we can form all its three boundary ellipses, each centered at one of its ``vertex''. 

\ul{Step (3.2): Common tangent line measurement.} At this point, we have acquired the contour functions of the three boundary ellipses. Yet, for the parts of the boundary of the ``Gaussian triangle'' that connect its different boundary ellipses, i.e., the common tangent lines between each pair of boundary ellipses shown by the dotted red lines in Fig.~\ref{fig:method}(c), we still haven't derived them. Here, to completely re-sketch the ``Gaussian triangle'' on the image plane, we discuss how we calculate the common tangent line between each pair of boundary ellipses. Notably, though this problem may sound complex at first glance, we highlight that, from the duality perspective, \textit{calculating the common tangent line of two (boundary) ellipses is equivalent to calculating the intersection of two dual (boundary) ellipses} \cite{richter2011perspectives}. 
Considering this, we show below that the common tangent line between each pair of boundary ellipses can be easily calculated in $O(1)$ time complexity through the following four stages:

\begin{figure*}[t]
\centering
\includegraphics[width=\textwidth]{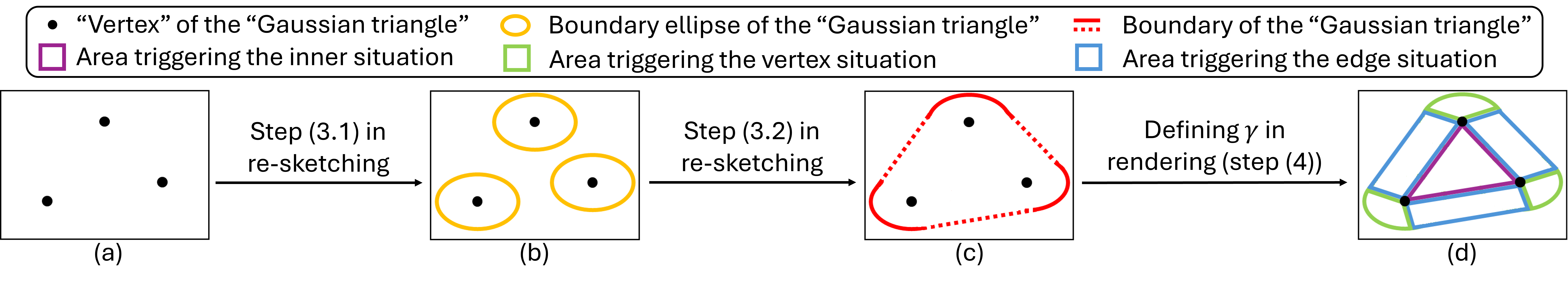}
\caption{Illustration of the re-sketching and rendering of the ``Gaussian triangle'' primitive on the image plane. Specifically, as shown from (a) to (c), via steps (3.1) and (3.2) introduced in Sec.~\ref{sec:4_1}, we first enable the boundary of the ``Gaussian triangle'' to be re-sketched on the image plane. After that, as shown in (d), with the fading parameter $\gamma$ re-defined for the ``Gaussian triangle'' over its different (sub-)areas, we enable the ``Gaussian triangle'' to properly join the $\alpha$-blending rendering process of Gaussian Splatting. (Best viewed in color.)}
\label{fig:method}
\end{figure*}

Firstly, for every boundary ellipse, we can easily find from Eq.~\ref{eq:method_3} that, its corresponding contour function $g^{ct}_i$ is actually in the form of:
\begin{equation}\label{eq:method_4}
\begin{aligned}
g^{ct}_i = a_i x^2 + b_i xy + c_i y^2 + d_i x + e_i y + f_i = 0
\end{aligned}
\end{equation}
where coefficients including $a_i$, $b_i$, $c_i$, $d_i$, $e_i$, and $f_i$ are all obtained through basic arithmetic operations from $\mu_i^{2D}$ and $\Sigma_i^{2D}$ (more details are provided in supplementary). 

Next, based on $g^{ct}_i$ in the form in Eq.~\ref{eq:method_4}, using the duality derivation in projective geometry \cite{richter2011perspectives}, we can build the dual boundary ellipse function $g^{dual}_i$ corresponding to $g^{ct}_i$ as:
\begin{equation}\label{eq:method_5}
\begin{aligned}
g^{dual}_i & = \big(-\frac{1}{4}(e_i)^2 + cf\big) x^2 + (\frac{1}{2}d_ie_i - b_if_i)xy \\
& ~~~+ \big(-\frac{1}{4}(d_i)^2 + a_if_i\big)y^2 + (-c_id_i + \frac{1}{2}b_ie_i)x \\
& ~~~+ (\frac{1}{2}b_ie_i - a_ie_i)y - \frac{1}{4}(b_i)^2 + a_ic_i = 0 \\
\end{aligned}
\end{equation}

After formulating $g^{dual}_i$, to find the common tangent line between, for example, the first and second boundary ellipses of a ``Gaussian triangle'', we need only determine the intersections of their corresponding dual boundary ellipses, by solving a system of two binary quadratic functions which holds closed-form solutions:
\begin{equation}\label{eq:method_6}
\begin{cases} 
    & g^{dual}_1 = 0 \\
    & g^{dual}_2 = 0
\end{cases}
\end{equation}

Finally, after finding the interactions of the two dual ellipses through solving Eq.~\ref{eq:method_6}, denoting $(x_s, y_s)$ one such derived interaction point, based on the duality property \cite{richter2011perspectives}, a common tangent line between the first and second boundary ellipses can then be derived simply as:
\begin{equation}\label{eq:method_7}
x_s\times x + y_s\times y + 1 = 0
\end{equation}

Notably, since Eq.~\ref{eq:method_6} is a system of quadratic functions, through Eq.~\ref{eq:method_6} and \ref{eq:method_7}, we could acquire multiple plausible common tangent lines. In this case, to properly represent the \textit{boundary} of the ``Gaussian triangle'' connecting its first and second boundary ellipses, as shown in Fig.~\ref{fig:method}, we select the common tangent line from the plausible ones that is farthest from the third ``vertex'' of the ``Gaussian triangle''. Denoting the chosen line as $x_s^c \times x + y_s^c \times y + 1 = 0$, to facilitate the later rendering process, we here further derive the point of tangency $t^1_{1,2}$ between the chosen line and the first boundary ellipse via solving the following equation system:
\begin{equation}\label{eq:method_12}
\begin{cases} 
    & x_s^c \times x + y_s^c \times y + 1 = 0 \\
    & g^{ct}_1 = 0
\end{cases}
\end{equation}
Here, since the line is a tangent line of the ellipse, we definitely can get one and only one real solution from solving Eq.~\ref{eq:method_12}. Similarly, we can also derive the point of tangency $t^2_{1,2}$ between the line and the second boundary ellipse.

With the above process repeated three times (i.e., performed over every pair of boundary ellipses), we can finally also acquire other points including $t^1_{1,3}$, $t^3_{1,3}$, $t^2_{2,3}$, and $t^3_{2,3}$. With these points, we prepare the ``Gaussian triangle'' on the image plane ready for rendering, as discussed below.

\textbf{Step (4): Rendering on the image plane.} To properly render a ``Gaussian triangle'' during the $\alpha$-blending rendering process of Gaussian Splatting, as shown in Eq.~\ref{eq:pre_2}, we need to formulate its color $c$ and its fading parameter $\gamma$. 
Among these parameters, for the color $c$ that is independent of the primitive's geometric shape, we use the same formulation for both the Gaussian ellipse and ``Gaussian triangle'' primitives, based on SH coefficient calculation. Yet, for $\gamma$ that is shape-relevant and originally defined in Eq.~\ref{eq:pre_2} based on the shape of Gaussian ellipse, we need to redefine $\gamma$ for it to align with the shape of ``Gaussian triangle''. Specifically, given a pixel $p$ of the image, depending on where the pixel $p$ overlaps with the ``Gaussian triangle'', over three different situations, we define its corresponding $\gamma$ in three different ways below.

To facilitate the later explanation, we first introduce some notations relevant to the current ``Gaussian triangle''. Specifically, we denote $\triangle$ the triangle formed by the three vertices $\{\mu_1^{2D}, \mu_2^{2D}, \mu_3^{2D}\}$. Besides, we denote 
$\square_{1,2}$ the quadrangle formed by the four vertices $\{\mu_1^{2D}, \mu_2^{2D}, t^2_{1,2}, t^1_{1,2} \}$, 
$\square_{1,3}$ the quadrangle formed by the four vertices $\{\mu_1^{2D}, \mu_3^{2D}, t^3_{1,3}, t^1_{1,3} \}$,
and $\square_{2,3}$ the quadrangle formed by the four vertices $\{\mu_2^{2D}, \mu_3^{2D}, t^3_{2,3}, t^2_{2,3} \}$. 

\ul{(i) The inner situation.} The first situation happens if $p$ lies inside the triangle $\triangle$ (i.e., in the purple area in Fig.~\ref{fig:method}(d)). Notably, no fading is expected in this inner area and we thus simply set $\gamma = \alpha$, where $\alpha$ is the opacity property of the ``Gaussian triangle'' itself. 

\ul{(ii) The vertex situation.} The second situation happens if $p$ lies in the green areas in Fig.~\ref{fig:method}(d) (i.e., neither in the triangle $\triangle$, nor in one of the quadrangles including $\square_{1,2}$, $\square_{1,3}$, and $\square_{2,3}$). In this case, denote the ``vertex'' of the ``Gaussian triangle'' that is nearest to $p$ the $j$-th ``vertex'' of the ``Gaussian triangle''. It can be observed from Fig.~\ref{fig:intro}(a) that, the fading of the ``Gaussian triangle'' at $p$ is then just equivalent to the fading of the Gaussian ellipse centered at $\mu^{2D}_j$. In light of this, similar to Eq.~\ref{eq:pre_2}, we define $\gamma$ under this situation as:
\begin{equation}\label{eq:method_9}
\gamma = \alpha e^{-\frac{1}{2}(p - \mu^{2D}_j)^T(\Sigma^{2D})^{-1}(p - \mu^{2D}_j))}
\end{equation}

\ul{(iii) The edge situation.} The third situation happens if $p$ lies in $\square_{1,2}$, $\square_{1,3}$, or $\square_{2,3}$ (i.e., the blue areas in Fig.~\ref{fig:method}(d)). In this case, to ensure the fading within the ``Gaussian triangle'' to transit smoothly and naturally across its different sub-areas (across different situations), we adopt a simple yet effective way to define the fading parameter $\gamma$ as follows.

Specifically, consider the case where $p$ lies in $\square_{1,2}$ first. Here, denote $e_1$ the edge of $\square_{1,2}$ connecting $\mu_1^{2D}$ and $t^1_{1,2}$, and $e_2$ the edge of $\square_{1,2}$ connecting $\mu_2^{2D}$ and $t^2_{1,2}$. Then let $p_{e_1}$ and $p_{e_2}$ respectively be points on $e_1$ and $e_2$, such that $p$ lies on the line segment between $p_{e_1}$ and $p_{e_2}$ and:
\begin{equation}\label{eq:method_10}
\begin{aligned}
\eta = &~\alpha e^{-\frac{1}{2}(p_{e_1} - \mu^{2D}_1)^T(\Sigma^{2D})^{-1}(p_{e_1} - \mu^{2D}_1))} \\
= &~\alpha e^{-\frac{1}{2}(p_{e_2} - \mu^{2D}_2)^T(\Sigma^{2D})^{-1}(p_{e_2} - \mu^{2D}_2))}
\end{aligned}
\end{equation}
We then set $\gamma = \eta$. Above we discuss how we measure $\gamma$ when $p$ lies in $\square_{1,2}$. The same procedure can be applied when $p$ lies in $\square_{1,3}$ or $\square_{2,3}$. 
Intuitively, this definition of the fading parameter $\gamma$ in the edge situation ensures that the value of $\gamma$ for pixel $p$ matches those assigned to the corresponding edge points $p_{e_1}$ and $p_{e_2}$. This thus preserves a smooth and coherent transition of the fading effect across the different sub-areas (i.e., across different situations) of the ``Gaussian triangle''. Meanwhile, it also yields a natural and gradual fading pattern within each individual quadrangle—$\square_{1,2}$, $\square_{1,3}$, and $\square_{2,3}$.

In summary, via the above four (bold) steps, given a ``Gaussian triangle'' in the 3D space, we can splat it onto the image plane, and corresponding render it through the $\alpha$-blending function with $\gamma$ redefined above. 

\textit{Splatting and rendering of the ``Gaussian line''.} Above we introduce how we perform splatting and rendering over the ``Gaussian triangle''. 
Here, we highlight that, the splatting and rendering of the ``Gaussian line'' can be achieved similarly, except in the following three places.
(1) Firstly, during primitive formulation (definition), for the ``Gaussian line'' with only two ``vertices'', we can discard $\mu_3$ and only define it with $\mu_1$ and $\mu_2$ to represent its ``vertices''. 
(2) Secondly, for the ``Gaussian line'', note that its two boundary ellipses are simultaneously connected by two common tangent lines. Hence, to properly re-sketch its boundary, after deriving the plausible common tangent lines between its two boundary ellipses through Eq.~\ref{eq:method_6} and \ref{eq:method_7}, from these candidates, we retain both common tangent lines that do not intersect each other in the middle, rather than retaining only one.
(3) Finally, since a line segment has no inner area, during the rendering of the ``Gaussian line'', we only need to consider the vertex and edge situations, but not the inner situation. 

With the rest process conducted similarly to the above process over the ``Gaussian triangle'', leveraging the compositional splatting strategy, we can also enable the splatting and rendering of the ``Gaussian line'' primitive. 
We also illustrate the splatting and rendering of the ``Gaussian line'' primitive in more detail in supplementary.

\subsection{Mixed Primitive-based Learning Procedure}
\label{sec:4_2}

Above, we discuss how we use a compositional splatting strategy within our framework to splat and render the ``Gaussian line'' and ``Gaussian triangle'' primitives.

However, the above strategy alone cannot fully support the learning procedure of Gaussian Splatting to be performed in a mixed-primitive-based manner. This is because, beyond splatting and rendering as core steps, the learning procedure of Gaussian Splatting also involves other steps. Some of these steps—such as initialization and pruning—are originally designed in the typical Gaussian Splatting pipeline under the assumption that only a single type of primitive is present.
As a result, directly incorporating these steps in their original form into our framework, with mixed types of primitives, can lead to incompatibilities, resulting in sub-optimal surface reconstruction performance. 
To tackle this problem, in our framework, we further adjust the typical Gaussian Splatting pipeline over its initialization and pruning steps. These adjustments then can better facilitate the mixed-primitive-based learning procedure in our MP-GS framework. Below, we describe these two steps in their adjusted forms one by one.

\noindent\textbf{The initialization step.} To ensure effective learning, it is important to initialize the learning procedure at a good starting point. 
The existing GS-based surface reconstruction methods \cite{huang20242d,dai2024high} smartly use the COLMAP point cloud as the prior information to formulate the starting point of its learning procedure. Specifically, they initialize a set of Gaussian ellipses each centered at a point in the COLMAP point cloud. However, though achieving good performance, this initialization strategy implicitly assumes every splatting primitive to be point-centered, which is not the case for ``Gaussian lines'' and ``Gaussian triangles''. This makes the usage of this initialization strategy in MP-GS improper. In light of this, we here aim to propose MP-GS with a new mixed primitive-based initialization strategy. 

To achieve this, assume that points in the COLMAP point cloud have been clustered into subsets, each \ul{containing 1-3 points} and meeting the criteria of \ul{close proximity} and \ul{similar colors}. For each subset, based on the number of points it has, we can then easily use it to initialize either the Gaussian ellipse with a single center point, the ``Gaussian line'' with two ``vertices'', or the ``Gaussian triangle'' with three ``vertices''. Note that here, we expect each subset of points to hold similar colors since as mentioned in Sec.~\ref{sec:4_1}, for each primitive, we only store it with a single copy of SH color coefficient $c_{SH}$ for parameter saving.

Considering the above, the challenge of equipping MP-GS with a proper initialization strategy now reduces to proposing a strategy that can cluster points in the COLMAP point cloud according to the above-specified criteria. Specifically, we find that, an effective way to perform such clustering involves the following two steps: 
(1) Firstly, we input all points from the COLMAP point cloud into a distance-based hierarchical clustering algorithm \cite{pitafi2023taxonomy} to generate a rooted clustering tree. Based on the algorithm, in this tree, each leaf node represents a single-point subset, while each non-leaf node represents the union of its child nodes' subsets. 
Meanwhile, for each node in the tree, the points in its corresponding subset are guaranteed to be of \ul{close proximity}.
(2) Next, we perform a breadth-first search (BFS) on the tree. During the BFS, a node is outputted if none of its ancestor nodes have been outputted, its subset \ul{contains 1-3 points}, and the points in the subset have \ul{similar colors}.
After completing the search in step (2), the collection of outputted tree nodes then allows us to cluster COLMAP points into subsets that meet the above-specified criteria. These point subsets can then be used to initialize the various types of primitives in our framework, providing a good starting point for its mixed-primitive-based learning procedure. We also include a pseudo-code algorithm about the above-introduced strategy in supplementary.

\noindent\textbf{The pruning step.} In the original Gaussian Splatting which uses only point-centered primitives, during the learning procedure, only primitive-level pruning is performed (e.g., when the opacity of a primitive is very low). Yet, in our framework which involves primitives with varying numbers of ``vertices'', we find that vertex-level pruning is sometimes also needed, e.g., in sub-areas of an object where ``Gaussian triangles'' are not necessary, and ``Gaussian lines'' and Gaussian ellipses with less number of ``vertices'' are enough. Hence, in our framework, whenever performing primitive-level pruning, we also perform vertex-level pruning. 

In specific, in our framework, based on the current shape of the primitive, we perform the following three types of vertex-level pruning: (1) Firstly, for a ``Gaussian triangle'' primitive, if its three ``vertices'' are all close to each other, this implies that the primitive can no longer need to retain all three of its ``vertices''. Hence, we prune its $\mu_2$ and $\mu_3$, and convert this primitive into a Gaussian ellipse centered at $\mu_1$. (2) Next, for a ``Gaussian triangle'' primitive, suppose its three ``vertices'' are not close to each other, but instead almost lie on the same line. In that case, we convert this ``Gaussian triangle'' into a ``Gaussian line''. (3) Lastly, for a ``Gaussian line'' primitive, if its two ``vertices'' are close to each other, similar to in (1), we prune its $\mu_2$ and reduce it to a Gaussian ellipse centered at $\mu_1$. Via the above, we enable the pruning of unnecessary ``vertices'' in the primitives during our framework's learning procedure. However, we emphasize that this pruning will not reduce our framework's surface representation to contain only point-centered Gaussian ellipses.
This is because, alongside the pruning step, the original densification mechanism of Gaussian Splatting is also integrated into our framework, by which our MP-GS framework can clone and split a steady stream of new ``Gaussian triangles'' and ``Gaussian lines'' if necessary.

In summary, by incorporating the above steps in their adjusted forms into our MP-GS framework, we ensure their compatibility with our framework, allowing our framework to seamlessly perform the learning procedure of Gaussian Splatting in a mixed-primitive-based manner.

\begin{table*}[t]
\caption{Results on DTU. We report Chamfer distance in millimeters. Lower Chamfer distance indicates better performance.}
\centering
\resizebox{\linewidth}{!}{
\begin{tabular}{l|ccccccccccccccc|c}
\hline
\multirow{2}{*}{\textbf{Method}} 
& \multicolumn{15}{c|}{Scene Index} & \multirow{2}{*}{Mean$\downarrow$} \\
\cline{2-16}
& 24 & 37 & 40 & 55 & 63 & 65 & 69 & 83 & 97 & 105 & 106 & 110 & 114 & 118 & 122 & \\
\hline 
NeRF \cite{mildenhall2021nerf} & 1.90 & 1.60 & 1.85 & 0.58 & 2.28 & 1.27 & 1.47 & 1.67 & 2.05 & 1.07 & 0.88 & 2.53 & 1.06 & 1.15 & 0.96 & 1.49 \\
VolSDF \cite{yariv2021volume} & 1.14 & 1.26 & 0.81 & 0.49 & 1.25 & 0.70 & 0.72 & 1.29 & 1.18 & 0.70 & 0.66 & 1.08 & 0.42 & 0.61 & 0.55 & 0.86 \\
NeuS \cite{wang2021neus} & 1.00 & 1.37 & 0.93 & 0.43 & 1.10 & 0.65 & 0.57 & 1.48 & 1.09 & 0.83 & 0.52 & 1.20 & 0.35 & 0.49 & 0.54 & 0.84 \\
Neuralangelo \cite{li2023neuralangelo} & 0.37 & 0.72 & 0.35 & 0.35 & 0.87 & 0.54 & 0.53 & 1.29 & 0.97 & 0.73 & 0.47 & 0.74 & 0.32 & 0.41 & 0.43 & 0.61 \\
\hline
3D-GS \cite{kerbl20233d} & 2.14 & 1.53 & 2.08 & 1.68 & 3.49 & 2.21 & 1.43 & 2.07 & 2.22 & 1.75 & 1.79 & 2.55 & 1.53 & 1.52 & 1.50 & 1.96 \\
SuGaR \cite{guedon2024sugar}  & 1.47 & 1.33 & 1.13 & 0.61 & 2.25 & 1.71 & 1.15 & 1.63 & 1.62 & 1.07 & 0.79 & 2.45 & 0.98 & 0.88 & 0.79 & 1.33\\
GaussianSurfels \cite{dai2024high} & 0.66 & 0.93 & 0.54 & 0.41 & 1.06 & 1.14 & 0.85 & 1.29 & 1.53 & 0.79 & 0.82 & 1.58 & 0.45 & 0.66 & 0.53 & 0.88 \\
2D-GS \cite{huang20242d} & 0.48 & 0.91 & 0.39 & 0.39 & 1.01 & 0.83 & 0.81 & 1.36 & 1.27 & 0.76 & 0.70 & 1.40 & 0.40 & 0.76 & 0.52 & 0.80 \\
GOF \cite{yu2024gaussian} & 0.50 & 0.82 & 0.37 & 0.37 & 1.12 & 0.74 & 0.73 & 1.18 & 1.29 & 0.68 & 0.77 & 0.90 & 0.42 & 0.66 & 0.49 & 0.74 \\
GS2Mesh \cite{wolf2024gs2mesh} & 0.59 & 0.79 & 0.70 & 0.38 & 0.78 & 1.00 & 0.69 & 1.25 & 0.96 & 0.59 & 0.50 & 0.68 & 0.37 & 0.50 & 0.46 & 0.68 \\
GeoFieldSplat \cite{Jiang_2025_CVPR} & 0.40 & 0.59 & 0.39 & 0.38 & 0.72 & 0.59 & 0.65 & 1.08 & 0.93 & 0.59 & 0.50 & 0.67 & 0.34 & 0.47 & 0.40 & 0.58 \\
\hline
Ours & 0.31 & 0.50 & 0.28 & 0.27 & 0.74 & 0.53 & 0.46 & 0.92 & 0.62 & 0.47 & 0.44 & 0.48 & 0.28 & 0.32 & 0.31 & \textbf{0.46} \\
\hline
\end{tabular}}
\label{tab:dtu}
\end{table*}

\begin{table*}[t]
\centering
\caption{Results on the Tanks\&Temples dataset. We report F1 score. Higher F1 score indicates better performance.}

\resizebox{0.7\linewidth}{!}{
\begin{tabular}{l|cccccc|c}
\hline
\multirow{2}{*}{\textbf{Method}} 
& \multicolumn{6}{c|}{Scene Name} & \multirow{2}{*}{Mean$\uparrow$}\\
\cline{2-7}
& Barn & Caterpillar & Courthouse & Ignatius & Meetingroom & Truck & \\
\hline
NeuS \cite{wang2021neus} & 0.29 & 0.29 & 0.17 & 0.83 & 0.24 & 0.45 & 0.38 \\
Geo-NeuS \cite{fu2022geo} & 0.33 & 0.26 & 0.12 & 0.72 & 0.20 & 0.45 & 0.35 \\
Neuralangelo \cite{li2023neuralangelo} & 0.70 & 0.36 & 0.28 & 0.89 & 0.32 & 0.48 & 0.50 \\
\hline
3D-GS \cite{kerbl20233d} & 0.13 & 0.08 & 0.09 & 0.04 & 0.01 & 0.19 & 0.09 \\
SuGaR \cite{guedon2024sugar} & 0.14 & 0.16 & 0.08 & 0.33 & 0.15 & 0.26 & 0.19 \\
2D-GS \cite{huang20242d} & 0.36 & 0.23 & 0.13 & 0.44 & 0.16 & 0.26 & 0.30 \\
GOF \cite{yu2024gaussian} & 0.51 & 0.41 & 0.28 & 0.68 & 0.28 & 0.59 & 0.46 \\
\hline
Ours & 0.66 & 0.44 & 0.21 & 0.81 & 0.33 & 0.66 & \textbf{0.52} \\ 
\hline
\end{tabular}}
\label{tab:tnt}
\end{table*}

\subsection{Overall Training and Testing}

In MP-GS, during training, we follow a similar process as the existing GS-based surface reconstruction methods \cite{dai2024high,chen2024pgsr}, except for the initialization and pruning steps, where we adopt the edited version introduced in Sec.~\ref{sec:4_2}. During testing, following \cite{huang20242d,chen2024pgsr}, we first render depth images from the learned surface representation. However, here, when splatting and rendering non-ellipical primitives, we depart from standard Gaussian Splatting and instead apply the compositional splatting strategy introduced in Sec.~\ref{sec:4_1}. After the splatting and rendering process, as in \cite{huang20242d,chen2024pgsr}, we use the rendered depth images to finally reconstruct the object surface via the TSDF algorithm \cite{newcombe2011kinectfusion}.

\section{Experiments}

To evaluate the surface reconstruction performance of our framework, we conduct experiments on 2 datasets including the DTU dataset and the Tanks\&Temples dataset. 

\noindent\textbf{DTU} \cite{jensen2014large} is a dataset popularly used in surface reconstruction. On this dataset, following existing surface reconstruction methods \cite{huang20242d,dai2024high,yu2024gaussian}, we evaluate our framework on a total of 15 scenes. Also following \cite{huang20242d,dai2024high,yu2024gaussian}, we use the Chamfer distance as the metric for evaluation. 

\noindent\textbf{Tanks\&Temples} \cite{knapitsch2017tanks} is another dataset that is commonly used in surface reconstruction. Following \cite{huang20242d,dai2024high}, we evaluate on 6 scenes on this dataset and use the F1 score as the evaluation metric.

\subsection{Implementation Details} 

We conduct our experiments on an RTX 6000 Ada GPU and develop our code based on \cite{kerbl20233d,chen2024pgsr,huang20242d}. During training, we use the same loss functions and loss weights as 2D-GS \cite{huang20242d}. Moreover, for fair comparison, we set the training iterations for all scenes to be 30,000. For the newly introduced parameters $\mu_1$, $\mu_2$, and $\mu_3$, we set their initial learning rates to 2e-4. Moreover, following \cite{huang20242d,yu2024gaussian}, we down-sample the input image with factor 2. 

During the initialization step, we consider points within a subset to have similar colors if their maximum pairwise color difference, measured using the L2 distance as described in \cite{compuphaseColourMetric}, is below $\omega_{color}$. In our framework, we set $\omega_{color}$ to 5. Moreover, during the pruning step of our framework, we regard a set of ``vertices'' as being close to one another if their maximum pairwise L2 distance is below $\omega_{dist}$, where $\omega_{dist}$ is set to 0.5 in our framework. 
Additionally, we determine whether the three ``vertices'' of a ``Gaussian triangle'' are nearly collinear by checking if the absolute value of the Pearson correlation coefficient computed from these ``vertices'' exceeds $\omega_{pear}$. In our framework, $\omega_{pear}$ is set to $0.9$.

Furthermore, during the densification process of our framework, when a ``Gaussian triangle'' is cloned or split into two new ``Gaussian triangles'', following how $\mu$ is assigned to each new Gaussian ellipse during the densification process of typical Gaussian Splatting \cite{kerbl20233d}, $\mu_1$ is assigned to each new ``Gaussian triangles'' in the same manner. 
Additionally, we assign both the new cloned/split ``Gaussian triangles'' with the same parameters $\mu_2$ and $\mu_3$ as the original ``Gaussian triangle''. 
Moreover, when a ``Gaussian line'' is cloned or split, we assign $\mu_1$ and $\mu_2$ to each new ``Gaussian line'' in the same way as how we assign $\mu_1$ and $\mu_2$ to each new ``Gaussian triangle'' as described above.

\subsection{Experimental Results}

\noindent\textbf{Quantitive Results.} In Tab.~\ref{tab:dtu} and Tab.~\ref{tab:tnt}, we compare our method with existing surface reconstruction methods. As shown, our method consistently achieves the best mean performance on both datasets, showing its effectiveness. 

\begin{figure*}[h]
\centering
\includegraphics[width=0.9\textwidth]{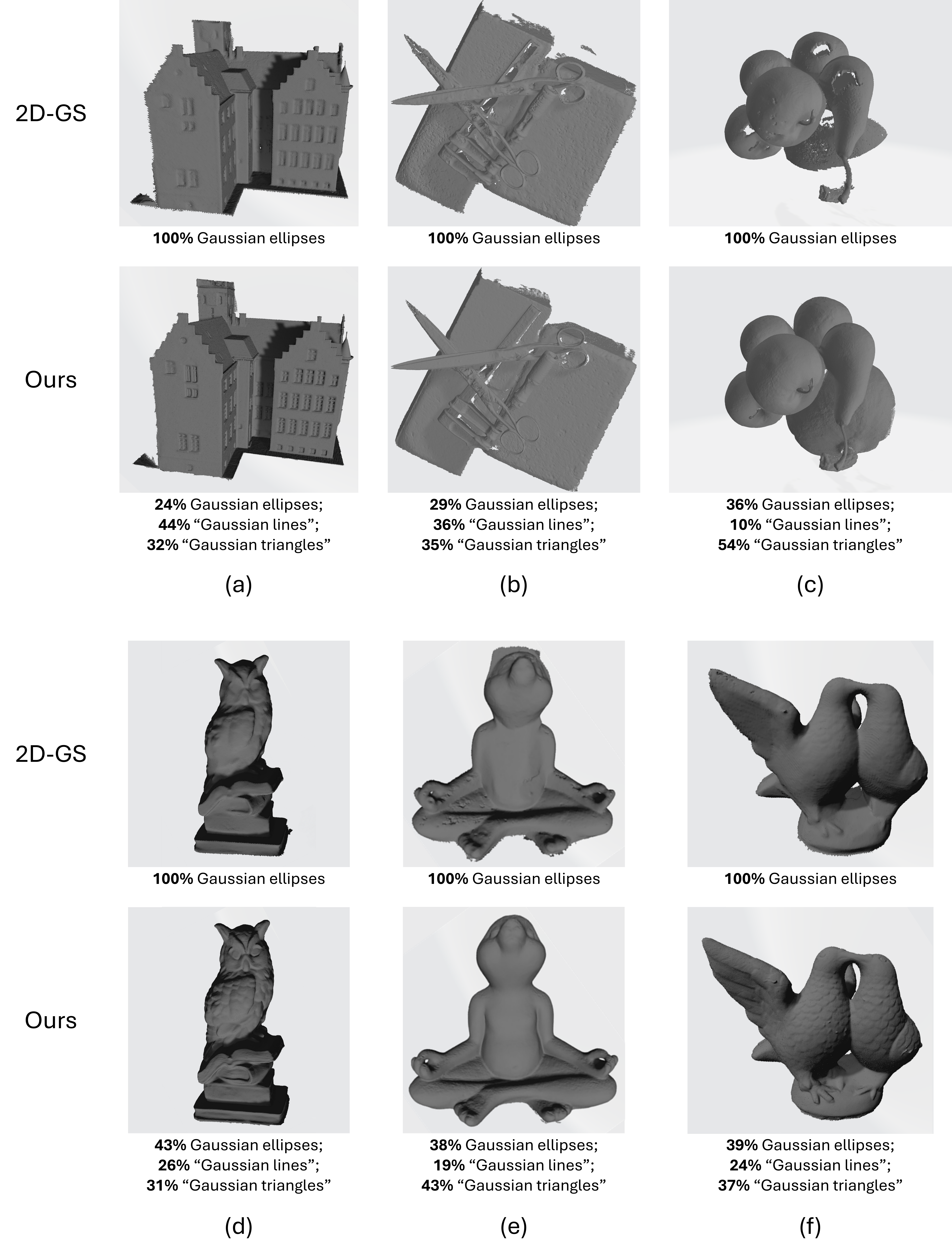}
\caption{Qualitative results of our MP-GS framework and the commonly-used GS-based surface reconstruction method 2D-GS \cite{huang20242d}. As shown, our framework based on mixed types of primitives achieves more accurate surface reconstruction than 2D-GS. \textbf{More qualitative results are in supplementary.}}
\label{fig:qualitative}
\end{figure*}

\noindent\textbf{Qualitative Results.} We also show some qualitative results on the DTU dataset in Fig.~\ref{fig:qualitative}. Additionally, in fig.~\ref{fig:qualitative}, we also illustrate the proportion of three types of primitives—Gaussian ellipses, ``Gaussian lines'', and ``Gaussian triangles''—within the learned representations of our MP-GS framework across different scenes. As shown, 2D-GS \cite{huang20242d}, a commonly used GS-based surface reconstruction method that relies solely on Gaussian ellipses with limited shape control \cite{pajarola2004confetti}, struggles to achieve high-quality object surface reconstruction. In contrast, our MP-GS framework introduces ``Gaussian lines'' to better represent long, thin structures, and ``Gaussian triangles'' to improve the reconstruction of relatively flat, smooth surfaces (due to their non-fading flat component in the middle, as shown in Fig.~\ref{fig:intro}). The adaptive use of these newly introduced primitives significantly enhances our framework’s reconstruction quality compared to 2D-GS. 

\begin{figure}[h]
\centering
\includegraphics[width=\columnwidth]{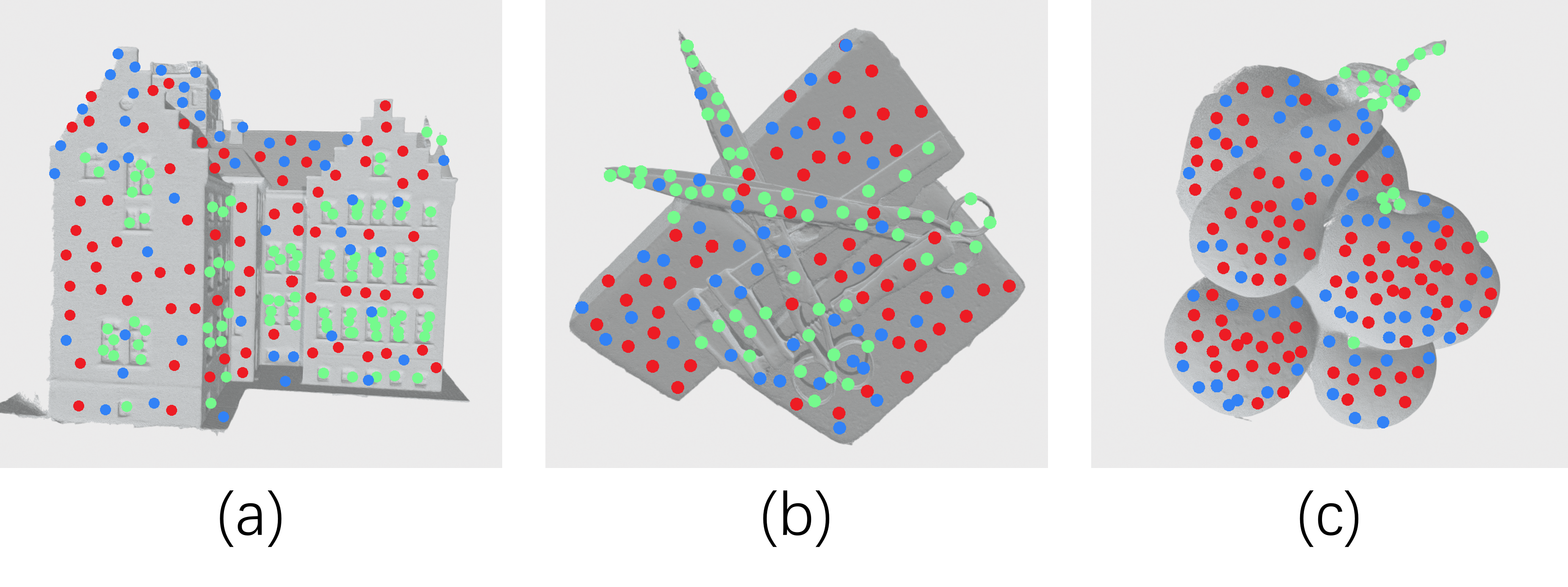}
\caption{Illustration on where ``Gaussian triangles'', ``Gaussian lines'', and Gaussian ellipses are distributed in our framework's reconstructions. \textcolor{red}{Red} dots show the positions of ``Gaussian triangles'', \textcolor{green}{green} dots show the positions of ``Gaussian lines'', \textcolor{blue}{blue} dots show the positions of Gaussian ellipses. For clarity, we down-sample before illustrating and display.}
\label{fig:supp_distribution}
\end{figure}

For example, in the window sub-areas of the scene shown in Fig.~\ref{fig:qualitative}(a) and the scissor blade sub-areas of the scene shown in Fig.~\ref{fig:qualitative}(b), which contain numerous long and thin structures, our MP-GS framework effectively utilizes a large proportion of ``Gaussian line'' primitives to represent these sub-areas, as shown in Fig.~\ref{fig:supp_distribution}(a) and (b), enabling more precise representation. Meanwhile, in the scene shown in Fig.~\ref{fig:qualitative}(c), 2D-GS exhibits holes in surface areas that should be relatively smooth and flat. In contrast, as shown in Fig.~\ref{fig:supp_distribution}(c), our MP-GS framework successfully reconstructs these regions using a large proportion of ``Gaussian triangles''. These results further demonstrate the effectiveness of our approach in improving surface reconstruction quality.

\subsection{Ablation Studies}

We conduct extensive ablation experiments on the DTU dataset, and report the Chamfer distance averaged over all the scenes. 

\noindent\textbf{Impact of the non-elliptical primitives.} In our framework, besides Gaussian ellipses, we additionally introduce Gaussian Splatting with non-elliptical primitives including ``Gaussian lines'' and ``Gaussian triangles''. To evaluate the efficacy of these two non-elliptical primitives, we test two variants. In the first variant (\textit{w/o ``Gaussian lines''}), we perform Gaussian Splatting in our framework only with Gaussian ellipses and ``Gaussian triangles'', while in the second variant (\textit{w/o ``Gaussian triangles''}), we involve our framework with only Gaussian ellipses and ``Gaussian lines''. As shown in Tab.~\ref{Tab:ablation_study_1}, our framework outperforms both these two variants, showing the importance of both types of non-elliptical primitives in our framework for accurately reconstructing object surfaces.

\begin{table}[ht]
\caption{Evaluation on the non-elliptical primitives.}
\centering
\resizebox{0.7\columnwidth}{!}
{
\small
\begin{tabular}{l|c}
\hline
Method & Chamfer distance $\downarrow$\\
\hline
w/o ``Gaussian lines'' & 0.61 \\
w/o ``Gaussian triangles'' & 0.64 \\
\hline
MP-GS & 0.46 \\
\hline
\end{tabular}}
\label{Tab:ablation_study_1}
\end{table}

\noindent\textbf{Impact of the proposed mixed primitive-based initialization strategy.} In our framework, to ensure compatibility between the initialization step in the learning procedure of Gaussian Splatting and the mixed-primitive-based nature of our framework, we propose a new mixed primitive-based initialization strategy (\textit{with proposed initialization strategy}). To validate its efficacy, we test a variant. In this variant (\textit{w/o proposed initialization strategy}), we do not apply the proposed initialization strategy. Instead, inspired by the typical Gaussian Splatting pipeline, we initialize a set of primitives, each centered at a point in the COLMAP point cloud and assigned a random primitive type. As shown in Tab.~\ref{Tab:ablation_study_2}, even without employing the proposed initialization strategy, our framework can still outperform the state-of-the-art method GeoFieldSplat \cite{Jiang_2025_CVPR}, while employing the mixed-primitive-based initialization strategy further improves our framework's performance.

\begin{table}[ht]
\caption{Evaluation on the proposed mixed primitive-based initialization strategy.}
\centering
\resizebox{0.85\columnwidth}{!}
{
\small
\begin{tabular}{l|c}
\hline
Method & Chamfer distance $\downarrow$\\
\hline
GeoFieldSplat \cite{Jiang_2025_CVPR} & 0.58 \\
\hline
w/o proposed initialization strategy & 0.56 \\
with proposed initialization strategy & 0.46 \\
\hline
\end{tabular}}
\label{Tab:ablation_study_2}
\end{table}

\noindent\textbf{Impact of the vertex pruning mechanism.} 
In our framework, to eliminate unnecessary ``vertices'' from the splatting primitives and thus refine the Gaussian Splatting representations into less messy ones, we propose a new vertex pruning mechanism (\textit{with vertex pruning}). To validate the efficacy of this mechanism, we test a variant (\textit{w/o vertex pruning}) in which we remove the vertex pruning mechanism from our framework. As shown in Tab.~\ref{Tab:supp_prune}, our framework involving the vertex pruning mechanism performs better than this variant. 
Additionally, we observe that, compared to this variant, our framework can reduce the average storage size required to store vertex coordinates by $24\%$.
The above shows the advantage of performing vertex-level pruning in addition to primitive-level pruning in our framework.

\begin{table}[ht]
\caption{Evaluation on the vertex pruning mechanism.}
\centering
\resizebox{0.65\columnwidth}{!}
{
\small
\begin{tabular}{l|c}
\hline
Method & Chamfer distance $\downarrow$ \\
\hline
w/o vertex pruning & 0.53\\
with vertex pruning & 0.46 \\
\hline
\end{tabular}}
\label{Tab:supp_prune}
\end{table}

\noindent\textbf{Time analysis.} Similar to existing GS-based surface reconstruction approaches \cite{dai2024high,huang20242d,yu2024gaussian}, we analyze the training time of our framework. Specifically, in Tab.~\ref{Tab:supp_training_time} we compare the training time of our framework with the existing NeRF-based surface reconstruction method Neuralangelo \cite{li2023neuralangelo}, as well as the existing commonly used GS-based surface reconstruction methods including 2D-GS \cite{huang20242d}, GaussianSurfels \cite{dai2024high}, and GOF \cite{yu2024gaussian}, on an RTX 6000 Ada GPU in terms of hours. As demonstrated, our MP-GS framework can achieve a competitive training time compared to existing GS-based surface reconstruction methods, while obtaining significantly better performance.

In addition to training time, here we also further analyze the rendering time of our framework, compared to 2D-GS which only uses Gaussian ellipses. We observe that, the rendering time (per image) of both our framework and 2D-GS is around 7ms on an RTX 6000 Ada GPU. This shows that, from the perspective of rendering efficiency, our method, whose main additional computation (common tangent line measurement) over typical Gaussian Splatting has closed-form solution, is also competitive compared to 2D-GS.

\begin{table}[ht]
\caption{Analysis of training time in terms of hours.}
\centering
\resizebox{0.85\columnwidth}{!}
{
\small
\begin{tabular}{l|cc}
\hline
Method & Chamfer distance $\downarrow$ & Training time \\
\hline
Neuralangelo \cite{li2023neuralangelo} & 0.61 & $>$12h \\
GaussianSurfels \cite{dai2024high} & 0.88 & 0.2h \\
2D-GS \cite{huang20242d} & 0.80 & 0.2h \\
GOF \cite{yu2024gaussian} & 0.74 & 1.0h \\
\hline
Ours & 0.46 & 0.3h \\
\hline
\end{tabular}}
\label{Tab:supp_training_time}
\end{table}

\noindent\textbf{Impact of the criteria in our proposed mixed primitive-based initialization strategy.} In our proposed mixed primitive-based initialization strategy, particularly during the point clustering process, besides close proximity, we require each subset of points to also meet the criterion of similar colors (\textit{close proximity + similar colors}). To validate this design choice, we test three variants. In the first variant (\textit{only close proximity}), we omit the additional criterion of similar colors during clustering. In the second variant (\textit{close proximity + similar normals}), instead of colors, we use surface normals—the other attribute in the COLMAP point cloud—as the additional criterion. In the third variant (\textit{close proximity + similar colors + similar normals}), we use both similar colors and similar normals as additional criteria. As shown in Tab.~\ref{Tab:supp_creteria}, regardless of the criteria used, our framework with the mixed primitive-based initialization strategy consistently achieves better performance than the variant \textit{w/o proposed initialization strategy} (defined in front of Tab.~\ref{Tab:ablation_study_2}). Meanwhile, we observe that incorporating color similarity alongside close proximity already enables the best performance among these variants, comparable to the variant that also incorporates similar normals. Thus, taking the framework complexity also into consideration, we adopt close proximity and similar colors as the clustering criteria in our mixed primitive-based initialization strategy.

\begin{table}[ht]
\caption{Evaluation on the criteria in our proposed mixed primitive-based initialization strategy.}
\centering
\resizebox{\columnwidth}{!}
{
\small
\begin{tabular}{l|c}
\hline
Method & Chamfer distance $\downarrow$\\
\hline
w/o proposed initialization strategy & 0.56\\
\hline
only close proximity & 0.50 \\
close proximity + similar normals & 0.48 \\
close proximity + similar colors  & 0.46 \\
close proximity + similar colors + similar normals & 0.46 \\
\hline
\end{tabular}}
\label{Tab:supp_creteria}
\end{table}

\noindent\textbf{Impact of the initial learning rate set to $\mu_1$, $\mu_2$, and $\mu_3$.} In our framework, during primitive formulation (definition), we introduce three new parameters including $\mu_1$, $\mu_2$, and $\mu_3$. For these parameters, in our experiments, we set their initial learning rates all to 2e-4 (i.e., $lr_{\mu} =$ 2e-4). Here, we also assess the other choices of $lr_{\mu}$ from 1e-4 to 1e-3, and report the results in Tab.~\ref{Tab:supp_lr}. As shown, with different choices of $lr_{\mu}$, our framework maintains consistent performance. This demonstrates the robustness of our framework to $lr_{\mu}$.

\begin{table}[ht]
\caption{Evaluation on the initial learning rate ($lr_{\mu}$) set to $\mu_1$, $\mu_2$, and $\mu_3$.}
\centering
\resizebox{0.5\columnwidth}{!}
{
\small
\begin{tabular}{l|c}
\hline
Method & Chamfer distance $\downarrow$\\
\hline
$lr_\mu$ = 1e-4 & 0.47 \\
$lr_\mu$ = 2e-4 & 0.46 \\
$lr_\mu$ = 5e-4 & 0.48 \\
$lr_\mu$ = 1e-3 & 0.49 \\
\hline
\end{tabular}}
\label{Tab:supp_lr}
\end{table}

\noindent\textbf{Impact of the hyperparameter $\omega_{color}$}. In the initialization step of our framework, we regard points in a subset to have similar colors if their maximum pairwise color difference, measured by L2 distance, is below $\omega_{color}$, where $\omega_{color}$ is set to 5 in our experiments. Here, we evaluate other choices of $\omega_{color}$ in Tab.~\ref{Tab:supp_thrc}. As shown, with different choice of $\omega_{color}$, the performance of our framework is consistent. This demonstrates the robustness of our framework to this hyperparameter.

\begin{table}[ht]
\caption{Evaluation on $\omega_{color}$.}
\centering
\resizebox{0.5\columnwidth}{!}
{
\small
\begin{tabular}{l|c}
\hline
Method & Chamfer distance $\downarrow$\\
\hline
$\omega_{color} = 1$ & 0.47 \\
$\omega_{color} = 5$ & 0.46 \\
$\omega_{color} = 10$ & 0.48 \\
$\omega_{color} = 20$ & 0.50 \\
\hline
\end{tabular}}
\label{Tab:supp_thrc}
\end{table}

\noindent\textbf{Impact of the hyperparameter $\omega_{dist}$}. In the pruning step of our framework, we consider a set of ``vertices'' of a primitive as being close to each other if their maximum pairwise L2 distance is below $\omega_{dist}$, where we set $\omega_{dist}$ to 0.5 in our experiments. We also evaluate other choices of $\omega_{dist}$ in Tab.~\ref{Tab:supp_thrd}. As shown, all variants ($\omega_{dist} = 0.1$, $\omega_{dist} = 0.5$, $\omega_{dist} = 1.0$, $\omega_{dist} = 2.0$) maintain a relatively consistent performance, demonstrating that our framework is fairly robust to the choice of $\omega_{dist}$, and does not require intensive tuning of this parameter.

\begin{table}[ht]
\caption{Evaluation on $\omega_{dist}$.}
\centering
\resizebox{0.55\columnwidth}{!}
{
\small
\begin{tabular}{l|c}
\hline
Method & Chamfer distance $\downarrow$\\
\hline
$\omega_{dist} = 0.1$ & 0.48\\
$\omega_{dist} = 0.5$ & 0.46 \\
$\omega_{dist} = 1.0$ & 0.47 \\
$\omega_{dist} = 2.0$ & 0.51 \\ 
\hline
\end{tabular}}
\label{Tab:supp_thrd}
\end{table}

\noindent\textbf{Impact of the hyperparameter $\omega_{pear}$}. Furthermore, in our framework, we regard the three ``vertices'' of a ``Gaussian triangle'' as being nearly collinear if the absolute value of the Pearson correlation coefficient measured from these three ``vertices'' exceeds $\omega_{pear}$. In our experiments, we set $\omega_{pear}$ to 0.9. In Tab.~\ref{Tab:supp_thrp}, we evaluate other choices of $\omega_{pear}$ as well. As shown, a compatible performance is achieved across different variants with different choices of $\omega_{pear}$. This shows that our framework is fairly insensitive to the choice of $\omega_{pear}$.

\begin{table}[ht]
\caption{Evaluation on $\omega_{pear}$.}
\centering
\resizebox{0.55\columnwidth}{!}
{
\small
\begin{tabular}{l|c}
\hline
Method & Chamfer distance $\downarrow$\\
\hline
$\omega_{pear} = 0.8$ & 0.49\\
$\omega_{pear} = 0.85$ & 0.46 \\
$\omega_{pear} = 0.9$ & 0.46 \\
$\omega_{pear} = 0.95$ & 0.47 \\ 
\hline
\end{tabular}}
\label{Tab:supp_thrp}
\end{table}

\section{Conclusion}

In this paper, we proposed a novel surface reconstruction framework MP-GS, which for the first time, enables Gaussian Splatting to perform surface reconstruction using a mix of elliptical and non-elliptical splatting primitives. Specifically, in MP-GS, we propose a novel compositional splatting strategy to enable the splatting and rendering of non-elliptical primitives. We also propose two other designs respectively over the initialization and pruning steps of Gaussian Splatting. Our framework achieves superior performance.

\bibliographystyle{IEEEtran}
{\tiny \bibliography{egbib}}


 




\vfill

\end{document}